\documentclass[11pt]{article}
\usepackage[final]{acl}
\usepackage{booktabs}
\usepackage{tabularx}
\usepackage{multirow}
\usepackage{longtable}
\usepackage{caption}
\usepackage{graphicx}
\usepackage{amsmath}
\usepackage{siunitx}
\usepackage{enumitem}
\usepackage[utf8]{inputenc}
\usepackage{authblk}
\usepackage{array}
\usepackage{footnote}
\usepackage{amssymb} 
\usepackage[table,dvipsnames]{xcolor}

\renewcommand{\arraystretch}{1.05}

\title{Who Is the Story About? \\Protagonist Entity Recognition in News}

\author{
Jorge Gabín\textsuperscript{1,2}\thanks{Corresponding author: jorge.gabin@udc.es}, M. Eduardo Ares\textsuperscript{1}, Javier Parapar\textsuperscript{2} \\
\\
\textsuperscript{1}Linknovate Science, A Coruña, Spain \\
\textsuperscript{2}IRLab, CITIC, Computer Science Department, University of A Coruña, Spain \\
\texttt{\{jorge.gabin,javier.parapar\}@udc.es, eduardo@linknovate.com}
}

\begin{document}
\maketitle

\begin{abstract}
News articles often reference numerous organizations, but traditional Named Entity Recognition (NER) treats all mentions equally, obscuring which entities genuinely drive the narrative. This limits downstream tasks that rely on understanding event salience, influence, or narrative focus. We introduce \textit{Protagonist Entity Recognition} (PER), a task that identifies the organizations that anchor a news story and shape its main developments. To validate PER, we compare the predictions of Large Language Models (LLMs) against annotations from four expert annotators over a gold corpus, establishing both inter-annotator consistency and human–LLM agreement. Leveraging these findings, we use state-of-the-art LLMs to automatically label large-scale news collections through NER-guided prompting, generating scalable, high-quality supervision. We then evaluate whether other LLMs, given reduced context and without explicit candidate guidance, can still infer the correct protagonists. Our results demonstrate that PER is a feasible and meaningful extension to narrative-centered information extraction, and that guided LLMs can approximate human judgments of narrative importance at scale.
\end{abstract}

\section{Introduction}

News articles are fundamentally narratives about actors and actions: they describe who did what, why it matters, and what consequences follow. While modern Information Extraction (IE) systems have made substantial progress in identifying entities and linking them to canonical knowledge bases \cite{grishman1997information}, they generally treat all entity mentions as equally important. In practice, however, some entities play a central narrative role, while others are merely contextual. For example, a financial report on a company's quarterly performance may reference competitors, regulators, and partners, yet only one or two organizations truly anchor the story. Traditional Named Entity Recognition (NER) captures all such mentions without distinguishing their narrative roles \cite{tjong-kim-sang-2003-conll}, leaving downstream applications such as media monitoring, event analysis, or knowledge-graph construction to filter large amounts of irrelevant information. What is often needed instead is an answer to a more interpretive question: \emph{who is the protagonist of this story?}

We refer to this challenge as \textit{Protagonist Entity Recognition (PER)} (Table~\ref{tab:per-example}). Unlike NER, which is primarily lexical and localized, PER is discourse-aware and interpretive. Determining a protagonist requires understanding how narrative emphasis is distributed across the document, integrating signals from headlines, leads, quotes, and event descriptions. Identifying protagonists therefore bridges the gap between surface-level mention detection and narrative-centered comprehension. While related work on entity and event salience \cite{dunietz-etal-2014-new, bhowmik-etal-2024-leveraging} aims to estimate importance, no existing benchmark directly evaluates which organizations are central to a news article’s narrative arc.

A major obstacle to progress is the absence of a task formalization and accompanying data. It is not clear, for instance, whether humans consistently agree on which entities in a news story are protagonists, nor how well current Large Language Models (LLMs) can approximate such judgments. To address this, we first conducted a controlled annotation study in which four expert annotators independently labeled the protagonist organizations in a set of 50 news articles. This gold-standard corpus enabled us to measure inter-annotator agreement and assess the reliability of the task itself. We then compared human annotations to LLM predictions, evaluating whether models can match the nuance of human narrative interpretation rather than simply identifying frequent or headline-mentioned entities.

Our findings indicate that, when appropriately guided \cite{ouyang-etal-2022-instructgpt}, LLMs achieve levels of agreement with human annotators comparable to the agreement levels among the annotators themselves. This result suggests that LLMs can serve as effective and reliable annotators for scaling the task beyond manually curated datasets. Building on this insight, we constructed a large-scale PER corpus by prompting state-of-the-art LLMs with NER-derived candidate entities, generating supervision signals that reflect narrative centrality rather than mere mention frequency. We further evaluated whether other LLMs, provided with substantially less context and without explicit candidate lists, can still identify protagonists purely from the document narrative.

\begin{table}[t]
\centering
\small
\begin{tabular}{p{7.4cm}}
\toprule
\textbf{Example Article Excerpt:}\\
\textit{``TechCorp announced record quarterly earnings on Tuesday, citing strong demand in Asian markets. The report also mentioned increased regulatory scrutiny from the European Commission and rising competition from GlobalSoft and InfoDynamics. Analysts expect TechCorp to expand further in the coming year.''}\\
\midrule
\textbf{Entities Recognized by NER:}\\
TechCorp, European Commission, GlobalSoft, InfoDynamics\\
\midrule
\textbf{Protagonist Entities:}\\
TechCorp\\
\bottomrule
\end{tabular}
\caption{Illustrative example showing that NER captures multiple entity mentions, while PER identifies which organization anchors the narrative.}
\label{tab:per-example}
\end{table}

This two-stage pipeline---manual validation followed by automatic expansion---contributes both a formal task definition and practical resources for studying narrative centrality in news. By demonstrating that PER is both learnable and scalable, we highlight its potential to improve a range of downstream applications that depend on understanding which actors truly matter in a story. More broadly, our work argues that information extraction should evolve from recognizing \emph{who is mentioned} to understanding \emph{who the story is about}, moving automated text analysis closer to genuine narrative comprehension.

\paragraph{Our Contribution.}
We build on these lines of research by (1) formalizing protagonist identification as a document-level discourse interpretation task, (2) establishing a human-annotated benchmark to assess reliability, and (3) evaluating whether contemporary LLMs can both replicate human judgments and scale labeling to larger corpora. In doing so, we connect human-centered annotation methodology with LLM-based corpus construction, and highlight when and how LLMs can approximate narrative understanding.

\section{Related Work}

\paragraph{Named Entity Recognition and Entity Salience.}
Traditional NER focuses on identifying and categorizing entity mentions \cite{tjong-kim-sang-2003-conll}, but it makes no inference about the narrative function of those entities. Entity salience modeling attempts to determine which entities are most central to a document’s meaning \cite{gamon2005identifying,liao2009simple,gillenwater-etal-2012-discovering}. Salience is typically estimated through features such as mention frequency, syntactic prominence, coreference chains, or discourse structure \cite{clark-manning-2015-entity}. However, salience does not necessarily imply protagonism: an entity may be frequently mentioned yet play a peripheral or reactive role. In contrast, \textit{Protagonist Entity Recognition} seeks to identify entities that drive the narrative, shape event progression, and define the story’s thematic focus.

\paragraph{Event Extraction and Semantic Role Labeling.}
Event extraction and semantic role labeling aim to determine \emph{who did what to whom} in specific event frames \cite{wei2019survey,marcus1993building}. While these approaches support structured representations of actions, they are event-centric rather than narrative-centric. A protagonist may appear across multiple events, episodes, or discourse segments, contributing to narrative continuity beyond any single predicate or event frame. This distinction aligns PER more closely with discourse-level analysis than with isolated event role identification.

\paragraph{Narrative Understanding and Summarization.}
Work on narrative structure, character modeling, and story summarization has highlighted the importance of identifying central actors in narrative texts \cite{elsner2012character,guan-etal-2020-knowledge}. However, most such work focuses on literary narratives or long-form stories, where protagonist roles are comparatively explicit. News narratives are compact, heterogeneous, and fact-driven, making protagonism less stylistically marked and more dependent on pragmatic, contextual, and discourse cues. PER thus extends narrative reasoning into the domain of real-world informational text.

\paragraph{LLMs for Annotation and Weak Supervision.}
Recent work has shown that LLMs can act as effective zero-shot or few-shot annotators for a wide range of linguistic tasks \cite{gilardi2023chatgpt,zhao2023survey}. LLMs have also been used to generate weak supervision \cite{meng2022generating}, enabling corpus expansion without manual labeling cost. However, only limited research has examined whether LLMs can reliably approximate \emph{interpretive} judgments---such as narrative centrality---that require global document reasoning rather than local classification. 
\section{Annotating Protagonists}

Establishing a reliable annotation framework is essential for studying PER. Since no prior dataset captures this notion, we first defined an annotation protocol and conducted a controlled human study to evaluate whether the concept of a \textit{protagonist organisation}---the entity that anchors the main narrative of a news article---can be consistently identified. Once the task proved interpretable and sufficiently reliable, we extended it to LLMs to examine whether they could emulate human judgement and scale corpus construction.

\subsection{Manual Annotation}

We sampled fifty articles from the Finer-ORD corpus \cite{finer-ord-2023}, which contains financial and organisational news. An off-the-shelf NER system \cite{tjong-kim-sang-2003-conll} was used to extract candidate organisational mentions; annotators could add missing entities when warranted. Four professional annotators, anonymised as A1--A4, were provided with concise guidelines framing protagonists as organisations whose actions, decisions, or state form the central focus of the article. Annotators were encouraged to treat the headline and lead paragraph as strong—but not definitive—signals, and to confirm whether the narrative focus was sustained throughout the body. Optional free-text rationales were included to support auditing of ambiguous cases.

\subsection{Automatic Annotation}

We then evaluated two prompting configurations of the same model family, LLaMA 3.3 70B. The zero-shot configuration, \textit{LLaMA-Base}, received the article and the list of candidate organisations and was asked to identify protagonists. The in-context configuration, \textit{LLaMA-ICL}, received the same instruction augmented with two manually annotated exemplars. This follows the in-context learning paradigm, where demonstration examples calibrate model decision boundaries \cite{brown2020language, dong2024survey}. Both configurations annotated the same 50-document sample used in the human study. Based on the comparative results, we used the in-context model to label the wider Finer-ORD corpus, including brief one-sentence justifications to enable efficient post-hoc auditing.

\subsection{Agreement Study}

Agreement was measured at the entity level using (i) average Jaccard similarity (mean per-document intersection-over-union), (ii) overall entity agreement (fraction of matching protagonist vs.\ non-protagonist labels), and (iii) Cohen’s $\kappa$ to correct for chance agreement under label imbalance \cite{cohen1960coefficient}. Table~\ref{tab:agreement-detailed} shows representative human–human, human–LLM, and model–model comparisons.

\begin{table}[t]
    \centering
    \scriptsize 
    \renewcommand{\arraystretch}{1.3} 
    \setlength{\tabcolsep}{7pt} 
    \caption{Pairwise agreement on the 50-document sample (411 total candidate entities).}
    \label{tab:agreement-detailed}
    \begin{tabular}{lccc}
        \toprule
        Pair & {Jaccard} & {Overall} & {$\kappa$} \\
        \midrule
        A1 -- A2 & 0.371 & 0.786 & 0.406 \\
        A1 -- A3 & 0.303 & 0.764 & 0.351 \\
        A1 -- A4 & 0.433 & 0.861 & 0.643 \\
        A2 -- A3 & 0.402 & 0.803 & 0.391 \\
        A2 -- A4 & 0.331 & 0.779 & 0.378 \\
        A3 -- A4 & 0.286 & 0.791 & 0.418 \\
        \arrayrulecolor{black!30}\midrule 
        A1 -- LLM-Base & 0.318 & 0.708 & 0.288 \\
        A2 -- LLM-Base & 0.467 & 0.757 & 0.366 \\
        A3 -- LLM-Base & 0.334 & 0.691 & 0.201 \\
        A4 -- LLM-Base & 0.309 & 0.701 & 0.265 \\
        \midrule
        A1 -- LLM-ICL & 0.365 & 0.752 & 0.328 \\
        A2 -- LLM-ICL & 0.494 & 0.810 & 0.427 \\
        A3 -- LLM-ICL & 0.345 & 0.754 & 0.267 \\
        A4 -- LLM-ICL & 0.328 & 0.749 & 0.314 \\
        \midrule\arrayrulecolor{black} 
        Base -- ICL & 0.714 & 0.878 & 0.689 \\
        \bottomrule
    \end{tabular}
\end{table}

Human–human comparisons show moderate to substantial agreement ($\kappa \approx 0.35$--$0.64$), confirming that protagonists are identifiable but require interpretive judgement. Overall agreement scores are inflated due to the large number of non-protagonist labels; $\kappa$ therefore provides a more informative indicator of shared decision criteria.

Human–LLM comparisons show that \textit{LLaMA-Base} over-selects protagonists, while \textit{LLaMA-ICL} calibrates its threshold more closely to human patterns. The two model variants also show high mutual consistency ($\kappa \approx 0.69$), indicating that exemplar conditioning stabilises model behaviour.

In summary, the agreement study demonstrates that (a) PER is a reproducible discourse-level task, (b) LLMs can approximate human judgement when exemplar-calibrated, and (c) scalable corpus construction is feasible with targeted human-in-the-loop auditing.

\section{Experiments}

\subsection{Experimental Setup}
We evaluate recent open large language models on the PER task using the test splits of two datasets: CoNLL-2003 \cite{tjong-kim-sang-2003-conll} and FiNER-ORD \cite{finer-ord-2023}. For both datasets, we use GPT-5 to generate the ``golden truth'' protagonist labels, given its strong reasoning capabilities and state-of-the-art performance among existing LLMs.

Each article provides a canonicalised list of candidate entities from which models must identify the central \emph{protagonists}. We compare two prompting configurations: (i) a base setup with a single task instruction, and (ii) an ICL setup extending the prompt with three exemplars representing a clear single-protagonist case, a co-protagonist case, and an ambiguous case—following standard in-context learning methodology \cite{brown2020language,dong2024survey}.

We test seven models spanning multiple architectures, sizes, and instruction-tuning regimes: LLaMA 4 16×17B, LLaMA 3.3 70B Instruct, LLaMA 3.1 70B, LLaMA 3.1 8B, Mistral-Nemo 12B, Gemma 3 27B, and Gemma 3 4B. All experiments use deterministic decoding (temperature = 0) to ensure reproducibility and isolate prompting effects.

Evaluation follows standard entity-level IE metrics: micro-F1, and macro-F1. Micro-F1 captures aggregate performance across all instances, while macro-F1 highlights robustness under class imbalance and document variation. Metrics are computed directly from model outputs without post-processing.

\subsection{Results and Discussion}
\begin{table*}[t]
    \centering
    \caption{Protagonist detection results across datasets. Values in parentheses show the effect of using ICL relative to the non-ICL baseline, with red indicating a decrease and green an increase.} 
    \scriptsize 
    \renewcommand{\arraystretch}{1.3} 
    \setlength{\tabcolsep}{7pt} 
    \label{tab:protagonist_results} 
    \begin{tabular}{lcllll} 
        \toprule 
        \multirow{2}{*}{\textbf{Model}} & \multirow{2}{*}{\textbf{ICL}} & \multicolumn{2}{c}{\textbf{CoNLL-2003}} & \multicolumn{2}{c}{\textbf{FiNER-ORD}} \\ 
        & & \multicolumn{1}{c}{Micro F1} & \multicolumn{1}{c}{Macro F1} & \multicolumn{1}{c}{Micro F1} & \multicolumn{1}{c}{Macro F1} \\ 
        \midrule 
        \multirow{2}{*}{Gemma-3 4B} & No & 0.297 & 0.315 & 0.308 & 0.351 \\ 
        & Yes & 0.330 {\tiny(\textcolor{Green}{+0.033})} & 0.300 {\tiny(\textcolor{red}{-0.015})} & 0.281 {\tiny(\textcolor{red}{-0.027})} & 0.270 {\tiny(\textcolor{red}{-0.081})} \\ 
        \arrayrulecolor{black!30}\midrule 
        \multirow{2}{*}{Gemma-3 27B} & No & 0.331 & 0.312 & 0.395 & 0.415 \\ 
        & Yes & 0.351 {\tiny(\textcolor{Green}{+0.020})} & 0.355 {\tiny(\textcolor{Green}{+0.043})} & 0.373 {\tiny(\textcolor{red}{-0.022})} & 0.382 {\tiny(\textcolor{red}{-0.033})} \\ 
        \midrule 
        \multirow{2}{*}{LLaMA-3.1 8B} & No & 0.337 & 0.213 & 0.341 & 0.330 \\ 
        & Yes & 0.320 {\tiny(\textcolor{red}{-0.017})} & 0.280 {\tiny(\textcolor{Green}{+0.067})} & 0.338 {\tiny(\textcolor{red}{-0.003})} & 0.333 {\tiny(\textcolor{Green}{+0.003})} \\
        \midrule 
        \multirow{2}{*}{LLaMA-3.1 70B} & No & 0.323 & 0.186 & 0.372 & 0.309 \\ 
        & Yes & 0.394 {\tiny(\textcolor{Green}{+0.071})} & 0.288 {\tiny(\textcolor{Green}{+0.102})} & 0.372 {\tiny(---)} & 0.362 {\tiny(\textcolor{Green}{+0.053})} \\ 
        \midrule 
        \multirow{2}{*}{LLaMA-3.3 70B} & No & 0.401 & 0.317 & 0.408 & 0.433 \\ 
        & Yes & 0.391 {\tiny(\textcolor{red}{-0.010})} & 0.363 {\tiny(\textcolor{Green}{+0.046})} & 0.414 {\tiny(\textcolor{Green}{+0.006})} & 0.426 {\tiny(\textcolor{red}{-0.007})} \\ 
        \midrule 
        \multirow{2}{*}{LLaMA-4 16x17B} & No & 0.225 & 0.105 & 0.434 & 0.371 \\
        & Yes & 0.350 {\tiny(\textcolor{Green}{+0.125})} & 0.207 {\tiny(\textcolor{Green}{+0.102})} & 0.467 {\tiny(\textcolor{Green}{+0.033})} & 0.446 {\tiny(\textcolor{Green}{+0.075})} \\ 
        \midrule\arrayrulecolor{black} 
        \multirow{2}{*}{Mistral-Nemo 12B} & No & 0.375 & 0.270 & 0.429 & 0.423 \\ 
        & Yes & 0.402 {\tiny(\textcolor{Green}{+0.027})} & 0.254 {\tiny(\textcolor{red}{-0.016})} & 0.429 {\tiny(---)} & 0.386 {\tiny(\textcolor{red}{-0.037})} \\ 
        \bottomrule 
    \end{tabular}
\end{table*}

Results are presented in Table~\ref{tab:protagonist_results}. For each model, we report performance under the base configuration and under ICL, and show the change relative to the base configuration.

The table reveals distinct operating regimes. Some models prioritise recall (e.g., LLaMA 3.3 Instruct and Gemma 3 27B), identifying many potential protagonists but at the cost of increased false positives. Others adopt more conservative selection policies (e.g., Mistral-Nemo 12B), improving precision but sometimes missing relevant entities.

ICL produces mixed effects. Larger models benefit most: LLaMA 4 shows substantial gains in both F1 and macro-F1 under ICL, indicating improved ability to detect protagonists across varied discourse cues. In contrast, smaller models occasionally experience performance degradation, likely due to exemplar miscalibration—ICL shifts their default inclusion threshold in ways that do not generalize across articles.

Macro-F1 deltas further show that ICL can improve consistency across documents for large models, while sometimes destabilizing smaller ones. This confirms that exemplar selection is a form of \emph{policy shaping}: exemplars must reflect the intended balance between precision and recall.

Overall, the results show that protagonist detection is tractable with current models, but is sensitive to calibration. Larger models benefit from exemplar conditioning, while smaller models require careful exemplar selection or light post-hoc filtering. This finding motivates an annotation workflow that pairs exemplar-guided large models with selective human auditing to scale PER reliably in practice.

\section{Conclusions}

This paper introduced and formalised the task of \textit{Protagonist Entity Recognition}, a discourse-level information extraction problem concerned with identifying which organisations constitute the central actors in a news narrative. In contrast to NER, which detects mentions without regard to narrative importance, protagonist detection targets \emph{narrative centrality}: determining which entities drive the events, decisions, and consequences that give the story its thematic focus.

We proposed an annotation framework and conducted a controlled human study across fifty news articles. The resulting agreement analysis showed that annotators converge meaningfully in identifying protagonists, indicating that the task is interpretable and reproducible. Moreover, a large model configuration demonstrated agreement with humans within the same range as inter-annotator variation, particularly when prompted with in-context exemplars. This finding supports the use of models as scalable, high-fidelity annotators.

Leveraging this insight, we automatically annotated the Finer-ORD corpus and evaluated a diverse set of recent models under matched settings. The results confirm that protagonist recognition is learnable: larger models yield robust performance, and exemplar-guided prompting provides measurable calibration benefits, especially in balancing inclusion thresholds. Smaller models, however, tend to overgeneralise protagonist roles, suggesting that their narrative reasoning remains brittle when signals are subtle or distributed across discourse segments.

Together, these contributions establish protagonist recognition as a tractable and meaningful extension of information extraction, one that moves beyond mention-level recognition toward entity-centric narrative understanding. The data, evaluation protocol, and baseline results provided here create a foundation for future work on discourse-focused entity modelling, media framing analysis, organisational influence tracking, and narrative structure learning. Continued progress will likely require expanding domain coverage, refining exemplar-based calibration strategies, and integrating protagonist signals into downstream applications such as summarisation, event forecasting, and knowledge graph enrichment.

\section*{Acknowledgements}
All the authors thank the annotators for their careful work and the team that assisted with prompt engineering and compute resources.
The authors from Linknovate Science gratefully acknowledge support from the grant \textit{IG408M – AXUDAS PARA O DESENVOLVEMENTO TECNOLÓXICO E A INNOVACIÓN MEDIANTE O USO DA INTELIXENCIA ARTIFICIAL (2025) – Liña A} for the project \textit{IA para Detección de Entidades Protagonistas (PEDIA)}.
\begin{figure}[h]
    \centering
    \includegraphics[width=\linewidth]{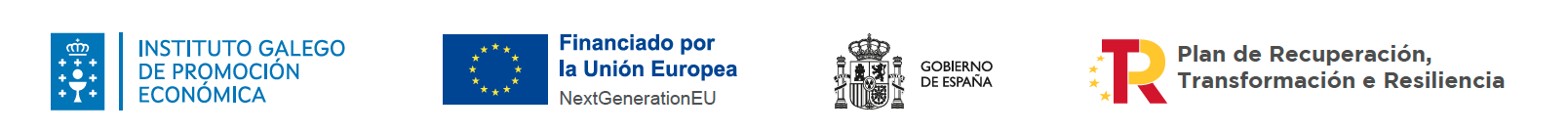}
    \label{fig:document}
\end{figure}
\section*{Limitations}

While the results are encouraging, several limitations remain. The annotated corpus is currently limited in size and domain, focusing on financial and organisational news. Although this domain provides clear narrative structures and well-defined organisational actors, generalising to political, environmental, local, or investigative reporting may require adapting both the annotation protocol and the decision criteria for narrative centrality. The task is inherently interpretive, and absolute agreement is unlikely even among humans; richer guidelines, adjudication procedures, and annotation discussions could further stabilise the notion of a protagonist across annotators and contexts.

Our large-scale annotations rely on a single LLM configuration, which may introduce systematic biases. Although exemplar conditioning improves alignment with human judgement, the choice of exemplars shapes the model’s protagonist selection policy, and different exemplar sets may produce different operating regimes. This highlights the need for systematic studies on exemplar diversity, representativeness, and robustness across models. Additionally, our evaluation assumes access to accurate candidate entity lists from upstream NER and canonicalisation; errors in these components propagate into protagonist predictions and may disproportionately affect entities with sparse or ambiguous references.

Finally, this work considers only English-language news. Narrative structures, discourse cues, and organisational naming conventions differ across languages and media ecosystems, making cross-lingual transfer non-trivial. Future work will explore multilingual protagonist detection, entity-type generalisation (e.g., persons and geopolitical entities), and the integration of protagonist signals into downstream applications such as event summarisation, influence tracing, and narrative graph construction. Beyond improving model accuracy, our aim is to position protagonist detection as a core component of broader news understanding pipelines, bridging the gap between entity recognition and narrative comprehension.

\bibliography{custom}

\newpage
\appendix
\section{Annotator guideline excerpt}
\begin{figure}[t]
    \centering
    \includegraphics[width=\linewidth]{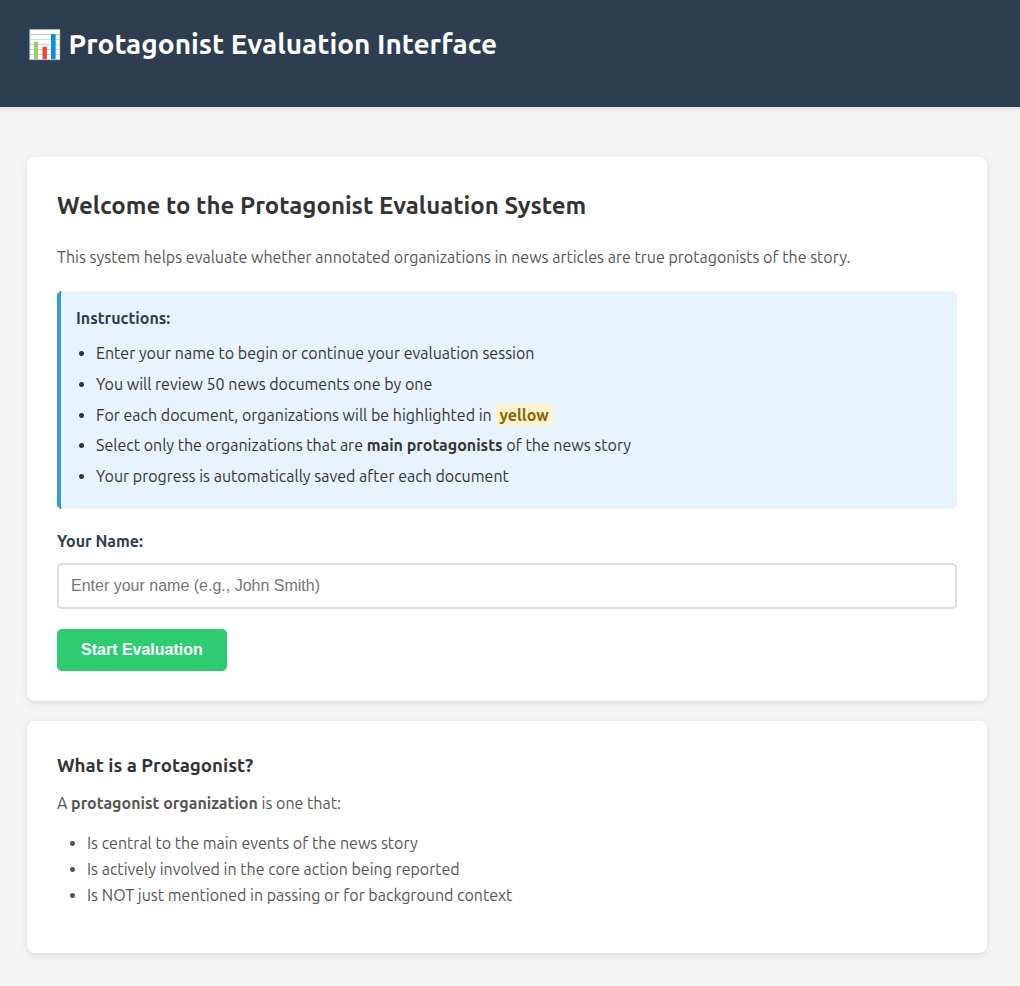}
    \caption{Initial screen showing the annotation guidelines and task overview. Annotators review examples before beginning the labelling process.}
    \label{fig:guidelines}
\end{figure}

Figure~\ref{fig:guidelines} presents the first screen of the annotation interface, where annotators review the task guidelines before starting. This section provides detailed instructions and practical examples illustrating how to identify the main organisational \emph{protagonists} of each news article. The interface ensures that annotators understand the distinction between central entities—those driving the story—and peripheral mentions, promoting consistency across annotations.

Once annotators confirm their readiness, they proceed to the main evaluation interface (Figure~\ref{fig:document}). Each document displays its unique identifier, the full news article with entity mentions highlighted, and a checklist of candidate organisations extracted via NER. Annotators must select those organisations that act as the main protagonists according to the contextual cues and narrative focus of the text. 

Progress indicators and validation buttons (e.g., “None are protagonists”, “Select All”, “Submit \& Continue”) guide annotators through the process and ensure that each decision is explicitly reviewed before submission. This design minimises omission errors and supports a transparent, auditable labelling workflow.

\begin{figure}[t]
    \centering
    \includegraphics[width=\linewidth]{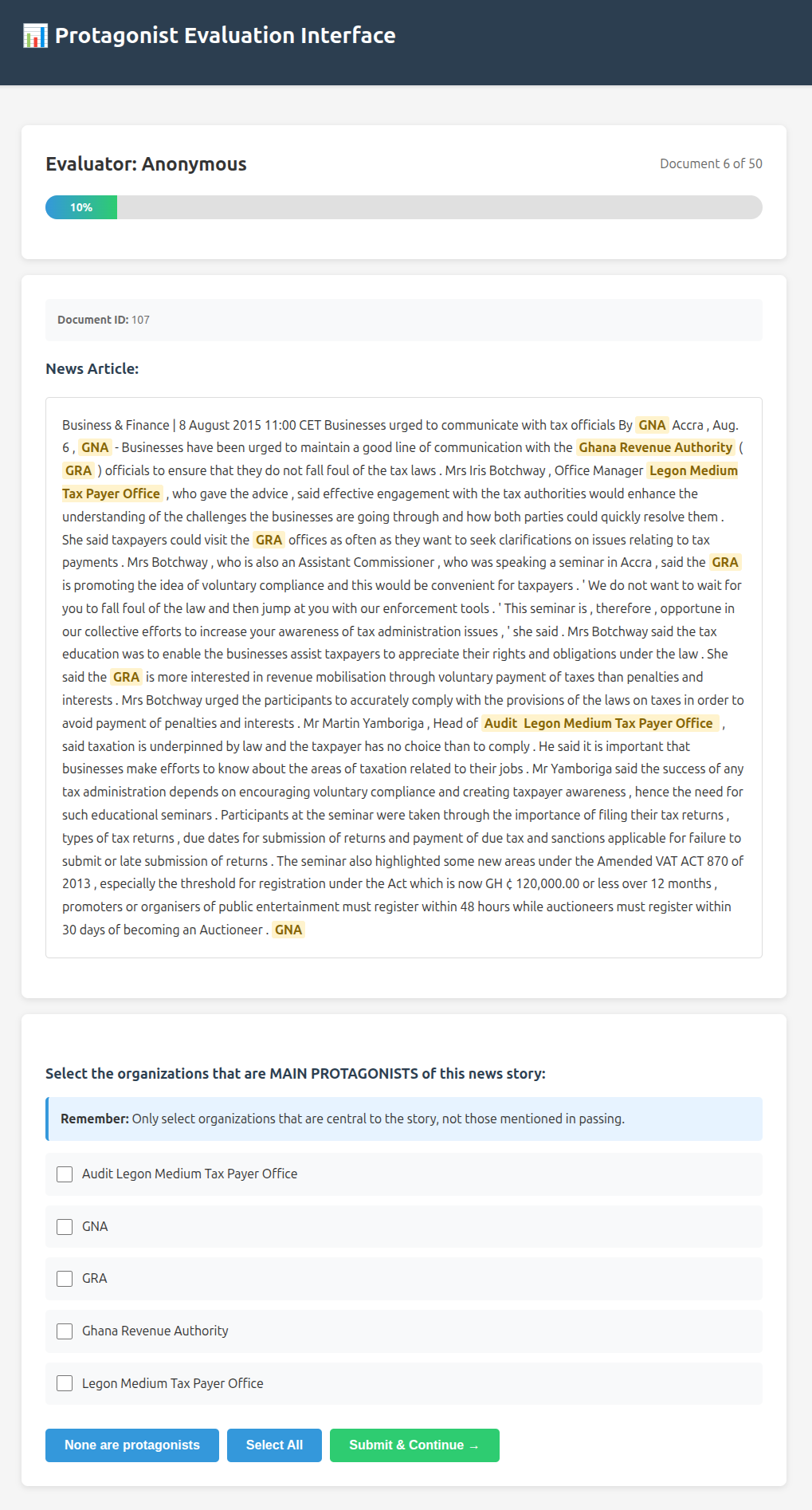}
    \caption{Main annotation interface displaying a news article, highlighted entity mentions, and candidate organisations for protagonist selection.}
    \label{fig:document}
\end{figure}

\end{document}